\documentclass[letterpaper, 10 pt, conference]{ieeeconf}  

\IEEEoverridecommandlockouts                              

\usepackage{graphicx}
\usepackage{booktabs}
\usepackage{subcaption}
\usepackage{multirow}
\usepackage{url}
\usepackage[dvipsnames]{xcolor}
\usepackage{pifont}
\newcommand{\cmark}{\ding{51}}
\newcommand{\xmark}{\ding{55}}

\newcommand{\revA}[1]{\textcolor{Black}{#1}}  
\newcommand{\revB}[1]{\textcolor{Black}{#1}}  
\newcommand{\revC}[1]{\textcolor{Black}{#1}}  
\newcommand{\revD}[1]{\textcolor{Black}{#1}}  

\usepackage{xcolor}

\title{\LARGE \bf
CroSTAta: Cross-State Transition Attention Transformer
\\ for Robotic Manipulation}

\author{Giovanni Minelli$^{1}$, Giulio Turrisi$^{1}$, Victor Barasuol$^{1}$, Claudio Semini$^{1}$ \\
        {\tt\small \{giovanni.minelli, giulio.turrisi, victor.barasuol, claudio.semini\}@iit.it}
\thanks{$^{1}$Istituto Italiano di Tecnologia, Genoa, Italy}%
}
\begin{document}

\maketitle
\thispagestyle{empty}
\pagestyle{empty}

\begin{abstract}
Learning robotic manipulation policies through supervised learning from demonstrations remains challenging when policies encounter execution variations not explicitly covered during training. While incorporating historical context through attention mechanisms can improve robustness, standard approaches process all past states in a sequence without explicitly modeling the temporal structure that demonstrations may include, such as failure and recovery patterns. We propose a \underline{Cro}ss-\underline{S}tate \underline{T}ransition \underline{At}tention Tr\underline{a}nsformer that employs a novel State Transition Attention (STA) mechanism to modulate standard attention weights based on learned state evolution patterns, enabling policies to better adapt their behavior based on execution history. Our approach combines this structured attention with temporal masking during training, where visual information is randomly removed from recent timesteps to encourage temporal reasoning from historical context. Evaluation in simulation shows that STA consistently outperforms standard attention approach and temporal modeling methods like TCN and LSTM networks, achieving more than 2× improvement over cross-attention on precision-critical tasks.
The source code and data can be accessed at the following link: \url{https://github.com/iit-DLSLab/croSTAta}


\end{abstract}

\section{Introduction}
Imitation learning (IL) has emerged as a promising paradigm for training robotic policies by leveraging expert demonstrations rather than learning a policy from scratch through extensive interaction with the environment \cite{drolet2024comparison}. The appeal of IL lies in its data efficiency and ability to leverage human expertise, making it particularly attractive for complex manipulation tasks \cite{ravichandar2020recent, chi2023diffusion, jiang2023vima, lee2024behavior, zare2024survey}. However, a fundamental limitation of IL approaches lies in the inherent dependence on the statistical distribution of training data, leading to brittle policies that struggle to handle situations not explicitly observed during training \cite{drolet2024comparison, de2019causal, spencer2021feedback}. This becomes even more relevant when deploying these models in unstructured and real-world scenarios where environmental conditions, object properties, or execution dynamics may differ from those observed in demonstrations \cite{de2019causal, wen2020fighting, mees2022matters, shao2025unifying}. %

\begin{figure}[t]
      \centering
        \includegraphics[scale=0.11]{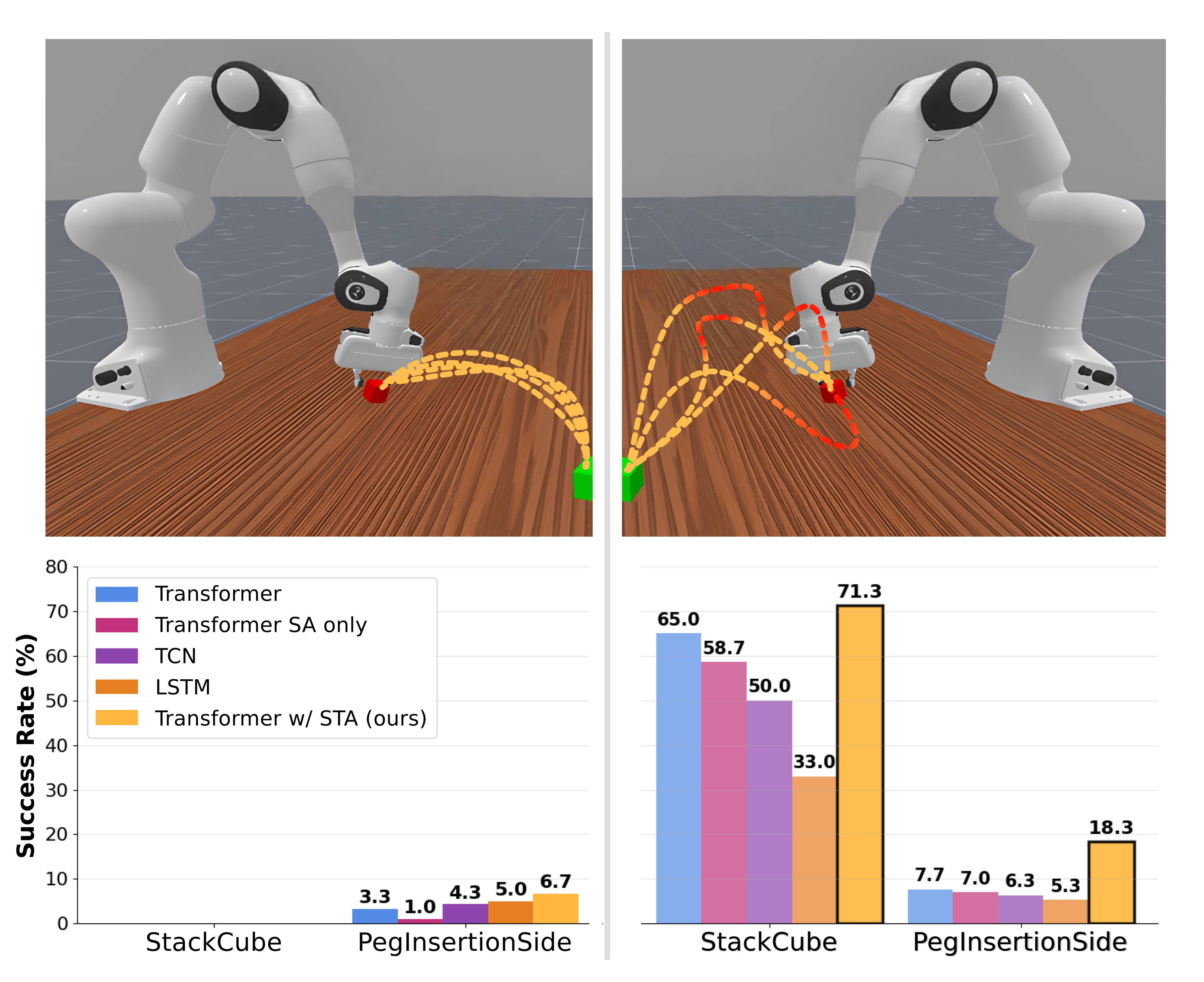}
        \caption{Performance comparison on simulated manipulation tasks when training
        with successful-only demonstrations (left) versus recovery-rich demonstrations (right).
        Our State Transition Attention (STA) mechanism shows particular effectiveness at
        exploiting temporal patterns in recovery-rich data, achieving superior performance
        compared to standard temporal modeling approaches.}
      \label{fig:teaser}
\end{figure}
   
To address this distributional shift problem, recent work has explored using suboptimal or noisy demonstration data, showing that sufficient diversity can sometimes outperform expert-only training, especially for long-horizon tasks \cite{kumar2022should, lin2025data}. This has motivated data augmentation approaches and automated demonstration generation systems \cite{jiang2023vima, james2020rlbench, mu2021maniskill, mandlekar2023mimicgen, gu2023maniskill2, garrett2024skillmimicgen, zhou2024autonomous} as methods to include diversity and failure/recovery trajectories in training data, thereby providing explicit examples of mistakes and corrections.

However, simply enriching demonstrations is not a scalable solution -- it would be extremely difficult to collect examples covering every possible failure scenario. This fundamental limitation highlights the need for approaches that can better leverage the underlying causal dependencies present in the data beyond straightforward sequence imitation. These dependencies span multiple aspects: logical dependencies where low-level actions depend on high-level plans; spatial dependencies where end-effector orientation depends on position; and crucially, temporal dependencies where future actions depend on past execution history. Alternative approaches based on planning \cite{feng2025reflective} and hierarchical \cite{li2025hamster} policy architectures address some of these challenges with promising results, but the fundamental question of how to extract dependency concepts from data and effectively model them in a policy remains central to achieving robust and adaptive behavior.

The temporal modeling challenge is particularly relevant because many robotic tasks are inherently non-Markovian: action selection often depends not only on the present, but also on past observations and actions \cite{lee2024behavior, mandlekar2022matters, zhao2023learning}. For example, manipulation scenarios where the robotic arm occludes critical scene information \cite{robocasa2024}, or multi-stage tasks where early steps inform later strategies \cite{chi2023diffusion, liu2025bidirectional}. In these cases, information used for decisions (e.g. speed, trajectory curvature, strategy) can fundamentally determine the execution of future actions. Yet learning long-context robotic policies through imitation learning remains challenging due to spurious correlations in extended observation histories \cite{villasevil2025learning}. 

Current sequence modeling approaches in robotics predominantly treat all temporal elements equally, learning relationships between past and present primarily through statistical co-occurrence of elements in demonstrated trajectories \cite{wen2023sequence}. While this approach has shown success in various domains, it may not optimally exploit the structured temporal dependencies in rich demonstrations, where specific past states inform corrective actions; thus, more targeted attention mechanisms could better capture these state transition relationships.

We propose a state transition attention mechanism that shifts attention-based temporal processing on how the past informs current action selection. Rather than extracting information from past timesteps and learning how to weight attention across the temporal dimension our approach directly learns to act based on state transition patterns. This allows policies to leverage historical context by matching current situations to learned temporal patterns during action selection. We evaluate our approach against standard mechanisms for temporal modeling and demonstrate its particular effectiveness in learning from recovery-rich demonstrations. Moreover, through analysis and ablations of the proposed method, we provide insights into how historical information is retrieved during execution phases, demonstrating that our structured attention mechanisms designed for state transition modeling can significantly enhance policy robustness in sequential decision-making.

The main contributions of this paper are:
\begin{itemize}
\item State Transition Attention (STA), a novel attention mechanism that modulates standard attention weights based on learned state evolution patterns, enabling explicit temporal reasoning over execution history in manipulation policies;
\item Empirical evaluation across four manipulation tasks demonstrating competitive performance of STA over standard attention approaches (up to 2× on precision-critical tasks) and over established temporal modeling baselines including TCN and LSTM, with ablation studies and attention pattern analysis providing insight into the mechanism's behavior.
\end{itemize}

\section{Related Work}
\begin{figure*}[ht]
  \centering
    \includegraphics[scale=0.185]{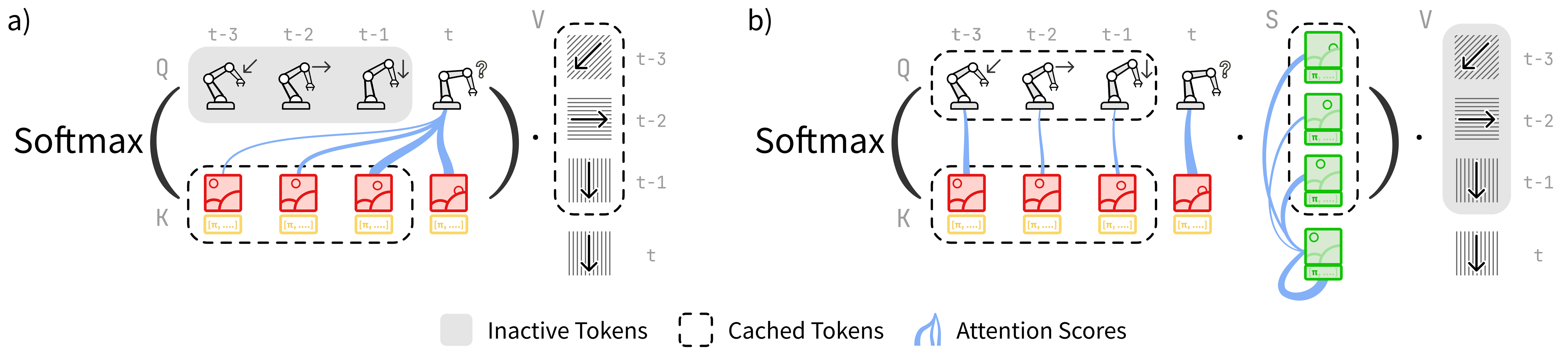}
    \caption{\revD{Graphical representation of how cross-attention works for computation at time $t$ within the Transformer architecture adopted}. Here, the query (Q) tokens represent joint values, while the keys (K), values (V) and state (S) tokens encode the overall system information. On the left, a) illustrates standard cross-attention and, on the right, b) depicts State Transition Attention.}
  \label{fig:attention-schema}
\end{figure*}

\subsection{Temporal Modeling and Attention Mechanisms}
The importance of historical context in robotic tasks has motivated
various temporal modeling approaches. Classical methods employ
Temporal Convolutional Networks (TCNs) \cite{lea2016temporalcn} and 
Long Short-Term Memory (LSTMs) \cite{hochreiter1997lstm} modules
for sequence processing, while more sophisticated approaches like
dynamic neural advection models incorporate temporal skip
connections to handle occlusion through context conditioning on
previous observations \cite{ebert2017selfsupervised}. The L-MAP
framework addresses temporal sequence modeling by learning
temporally extended macro-actions for scalable decision-making
\cite{luo2025scalable}. 
Recent transformer-based \cite{vaswani2017attention} approaches
rely on sequential modeling capabilities for temporal decision-making
in robotics. Decision Transformers and Trajectory Transformers apply
sequence modeling within reinforcement learning frameworks, modeling
task execution as autoregressive sequence prediction \cite{chen2021dt, janner2021offline}.
In contrast, imitation learning approaches employ sequence modeling
through different strategies: methods like VIMA demonstrate
capabilities for long-horizon tasks and generalization \cite{jiang2023vima},
while others predict sequences of actions in chunks \cite{chi2023diffusion, zhao2023learning, black2024pi0, nvidia2025gr00tn1openfoundation}. In-context learning approaches \cite{fu2024incontext}
leverage Transformers' sequential nature by enriching input
sequences with example trajectories that attention mechanisms can
retrieve from. 
Standard attention mechanisms have proven excellent for sequence
modeling applications, demonstrating strong noise resistance and
long-sequence information retrieval capabilities \cite{wang2024needle}.
Nonetheless, they primarily learn state relationships through
statistical co-occurrence rather than explicitly modeling the
implications of state evolution. This limitation can lead to
failures when execution variations or world state interpretations
diverge from training distributions. Recent work has begun
addressing this limitation through bidirectional approaches that
enforce coherence between past and future predictions \cite{villasevil2025learning, liu2025bidirectional},
yet the fundamental challenge of effectively leveraging structured
temporal dependencies in demonstrations remains
largely unexplored.

\section{Method}
\revD{Effective utilization of historical information becomes critical in robotic manipulation tasks where current observations alone may be insufficient for reliable action selection, whether due to ambiguous scene configurations, execution imprecisions, or dynamic environments where conditions evolve in ways not fully captured at a single timestep}. To achieve this, we need both a mechanism capable of modeling how past states relate to current conditions and training data with informative temporal patterns that can aid learning. Our preliminary tests demonstrate that recovery-rich demonstrations -- which contain explicit failure-to-recovery patterns -- consistently improve performance across different temporal modeling methods compared to training on successful trajectories alone (Fig. \ref{fig:teaser}). This suggests that such demonstrations provide the necessary informative temporal structures for learning robust policies. Building on this insight, we propose a state transition attention mechanism designed to exploit these temporal patterns.

\subsection{State Transition Attention Mechanism}
We propose a modification to standard attention mechanisms that focuses on state transition patterns rather than individual past states. The key intuition is that the relational patterns most relevant for current decision-making, emerge from understanding how states evolve over time, particularly, observing direct relationships between subsequent states.

Following an architectural approach similar to \cite{nvidia2025gr00tn1openfoundation}, we use an encoder-decoder with cross-attention to relate decoder actions to encoder state information e.g., in a robotic context, joint movements (actions) to visual/sensor inputs (states). Standard cross-attention mechanisms in this setup learn to relate current actions to all present and past states through learned linear projections. These projections must learn representations valid across all temporal distances, while positional embeddings distinguish different timesteps and softmax operations weight the relevance of historical events. This approach places a significant burden on the attention mechanism to both learn appropriate representations and correctly dampen irrelevant historical information through softmax normalization.

\revD{Instead, we shift the computational focus toward interpreting state transition patterns, by using relationships between current and past states to reproject attention scores relative to the different timesteps (i.e., between past action tokens and past state tokens)}. In (\ref{eq:cross-att}), we formalize standard cross-attention applied to a temporal sequence, and in (\ref{eq:st-att}) we formalize our proposed State Transition Attention (STA) mechanism:

\begin{equation}
Softmax(\frac{Q_{t}K^T_{t-k:t}}{\sqrt{d_K}})V_{t-k:t}
\label{eq:cross-att}
\end{equation}
\begin{equation}
Softmax(\frac{diag(Q_{t-k:t}K^T_{t-k:t})(S_{t-k:t}S^{\revD{T}}_t)}{\sqrt{d_K  d_S k}})V_t
\label{eq:st-att}
\end{equation}
where $Q, K, V, S$ are independent linear neural network projections of decoder ($Q$) and encoder ($K, V, S$) tokens; subscripts span the sequence from time $t$ back $k$ steps; and $d_K$ and $d_S$ refer to the dimensions of the respective projection matrices. \revA{The state transition projection $S$ learns to identify which historical states are most relevant given the current state, creating \revB{transition-aware attention values that are then multiplied with the diagonal elements of $QK^T$, i.e., $Q_{t-k}K^T_{t-k}, \dots, Q_{t}K^T_{t}$, deliberately \revD{decoupling per-timestep action-state alignment from cross-temporal relevance}, which is instead captured by the state transition projection $S$}.} \revA{The normalization factor follows standard scaled dot-product attention practices, accounting for the dimensionality of components \cite{vaswani2017attention}}. \revB{Crucially, the softmax operation in STA is applied only over current timestep tokens ($n$ tokens) rather than across the entire history ($(k+1) \cdot n$ tokens), reducing the computational cost of the exponential operations. However, this is offset by the additional projection $S$ and dot product $S_{t-k:t}S^T_t$ of $O(k n^2 d_S)$, resulting in similar overall computational cost while providing different representational capabilities. At inference, both approaches benefit from caching strategies \cite{huggingface_kvcache_2025}, with STA caching $Q$, $K$, and $S$ projections (with $Q$ and $K$ potentially cached in form of attention scores since they don't depend on future steps), instead of $K$ and $V$.} To maintain temporal awareness while focusing on state evolution patterns, \revA{we add element-wise learned absolute positional embeddings into the state transition projection output $S$}. A graphical comparison between attention strategies is provided in Fig. \ref{fig:attention-schema}.

\subsection{Architecture Design}
Building upon the STA component, our decoder uses standard Transformer blocks with STA cross-attention, self-attention, and feed-forward layers. It processes \textit{input tokens} representing coordinated joint actions. To better exploit relational patterns through self-attention mechanisms, we use an \textit{input token} per single joint action, initialized with proprioceptive information and absolute positional embeddings \cite{vaswani2017attention} to establish intra-relationships between them. Output tokens are individually processed through an MLP to produce the target action for each joint, then executed through an underlying PD controller. We refer to this architecture as STA Transformer for brevity in the remainder of the paper.

\begin{figure}[t]
      \centering
        \includegraphics[scale=0.165]{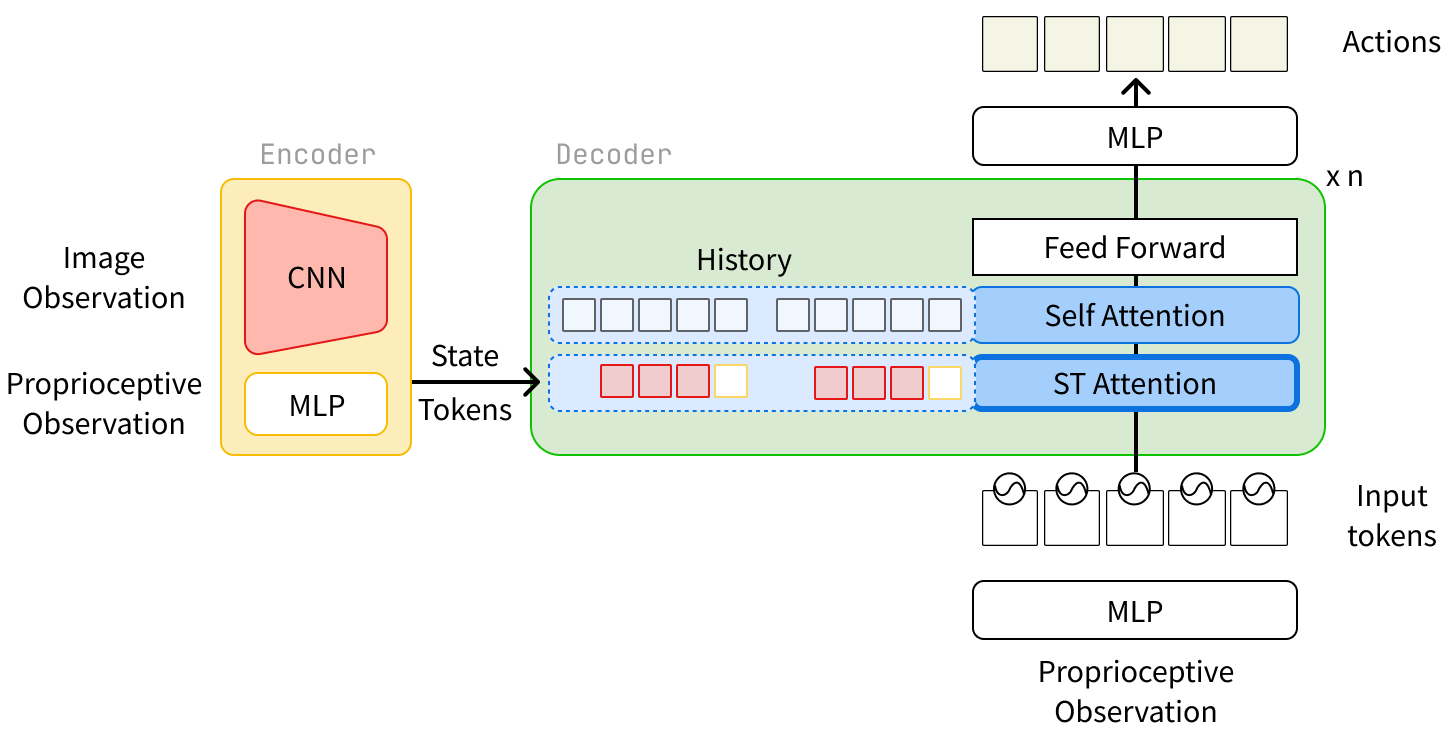}
        \caption{Architecture overview of our proposed Transformer with
        STA. The encoder processes visual observations
        through CNN and proprioceptive data through MLP to generate \textit{state tokens}.
        The decoder employs standard self-attention for \textit{input token} interactions
        (white squares) and our novel STA module as cross-attention with current
        and historical \textit{state tokens} (colored squares).
        Both decoder input's tokens and encoder state-related
        tokens are being cached for reuse in later steps.}
      \label{fig:net}
\end{figure}

The encoder deals with world state information processed through a convolutional neural network (CNN) and an MLP network to handle visual and proprioceptive inputs, respectively. The concatenated output of these modules represents the \textit{state tokens} for the decoder's cross-attention layers. The overall architecture allows to process the evolution of the world state (through STA cross-attention) and to model the robot's internal kinematics (through self-attention), providing both stages with tokens of current and past timesteps. A complete overview of the architecture is provided in Fig. \ref{fig:net}.

\subsection{Training Strategy with Temporal Masking}
For training, we sample sequences of 16 timesteps where each one uses the previous steps as contextual history, while future information remains masked. We optimize the model using mean squared error loss between predicted and ground-truth actions. \revD{Additionally, to incentivize information retrieval from historical context and enhance learning of temporal dependencies, we propose a temporal masking strategy applied to visual inputs provided to the encoder. \revA{All exteroceptive information is removed for $k$ consecutive timesteps (excluding the first/oldest), where $k$ is randomly sampled from [2, $L$/2] with $L$=16 being the total sequence length}. This masking strategy serves a dual purpose: it prevents the model from developing over-reliance on current visual information while simultaneously encouraging the development of robust temporal reasoning capabilities, forcing the model to rely on historical context for decision-making}.

\section{Evaluation}
\begin{figure*}[htbp]
    \centering
    \includegraphics[scale=0.18]{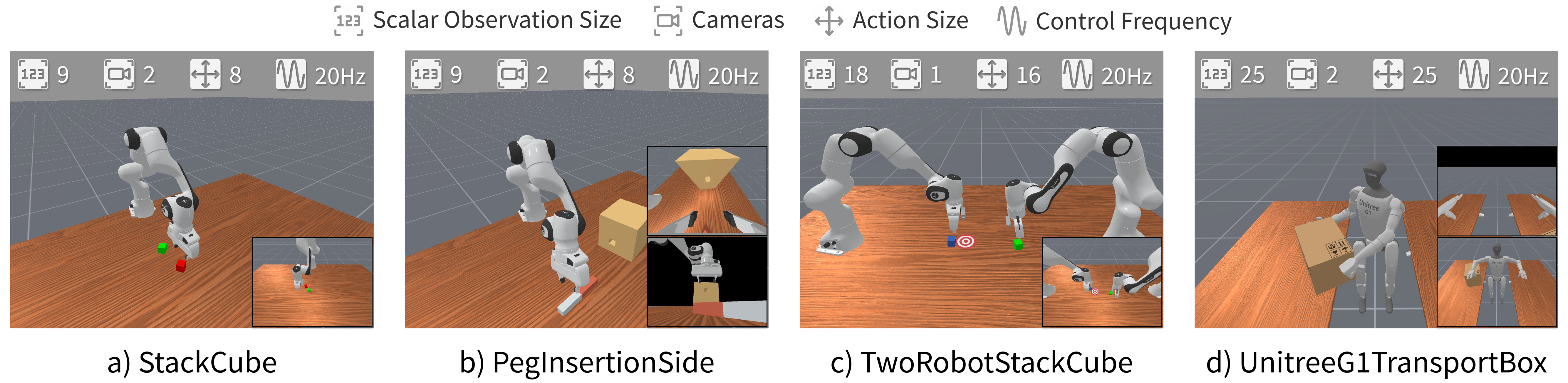}
    \caption{ManiSkill manipulation tasks used for evaluation. (a) \textit{StackCube}: Single-arm manipulation requiring coordinated grasping and placement; (b) \textit{PegInsertionSide}: Precision insertion task demanding correct orientation and alignment of the peg with the box hole slot; (c) \textit{TwoRobotStackCube}: Bimanual coordination task for collaborative cube stacking in a target location; (d) \textit{UnitreeG1TransportBox}: Multi-joint coordination task involving arm and torso coordination in a humanoid robot to transport a box across the workspace.
    }
    \label{fig:maniskill-tasks}
\end{figure*}

\subsection{Evaluation Setup}
We evaluate STA mechanism on four ManiSkill tasks \cite{mu2021maniskill} chosen for distinct dynamics and failure modes (Fig. \ref{fig:maniskill-tasks}): \textit{StackCube} and \textit{PegInsertionSide} require precise manipulation; \textit{TwoRobotStackCube} demands coordinated manipulation of two objects where failure in either subtask compromises the task; and \textit{UnitreeG1TransportBox} requires synchronized torso rotation and arm movements. \revC{These tasks feature short horizons (less than 5 seconds) but remain failure-prone due to their demanding execution requirements, with task configurations that can cause camera occlusions interfering with visual information, e.g., during the manipulation of pegs in \textit{PegInsertionSide} and the mutual interference of arms in \textit{TwoRobotStackCube}, making them ideal testbeds for evaluating temporal reasoning capabilities under randomized inference conditions.} \revD{Each task uses one or two cameras to capture visual information (see Fig. \ref{fig:maniskill-tasks}), while proprioceptive state information consists exclusively of joint position values}. Action prediction consists of joint position delta values across all evaluated methods.

\subsection{Data Collection}
We collect demonstrations containing artificially induced failure sequences followed by natural recovery behaviors, following a DAgger-like methodology \cite{drolet2024comparison}. Data is generated using policies trained with privileged information via PPO, with noise-robustness explicitly enforced through output noise injection during training. At collection time, random perturbations to the state representation of world elements for $n$ consecutive steps force the policy to perform suboptimal actions (e.g., misperceiving object positions leading to failed grasps or collisions), after which it naturally recover toward the true target, yielding trajectories with explicit failure-correction patterns.
Crucially, since failures are artificially induced, we label the collected trajectory steps to train exclusively on noise-free action predictions. Although predictions for failure-induced steps do not contribute to the training loss, the visited states and generated trajectory segments remain key components of the sequential network input, providing rich temporal context for learning state transition patterns. We generated 1000 episodes per task for training.

\subsection{Baseline Methods}
We compare our STA Transformer against five baselines: a Transformer with standard cross-attention over state sequences, a Transformer that does not take past historical context into account, \revD{a self-attention-only Transformer processing state and input tokens together}, and established temporal modeling approaches including TCN \cite{lea2016temporalcn} and LSTM \cite{hochreiter1997lstm} networks. All networks use 4 layers and hidden size 512. We equipped TCN and LSTM networks with additional final feed-forward networks to predilige a fair performance comparison given the parameter count differences with transformer-based methods. The encoder structures remain comparable across all methods, with a hidden representation of 512. Additional implementation details and training hyperparameters are provided in Table \ref{tab:params-networks}.

\subsection{Performance Evaluation}

   \begin{figure}[t]
      \centering
        \includegraphics[scale=0.17]{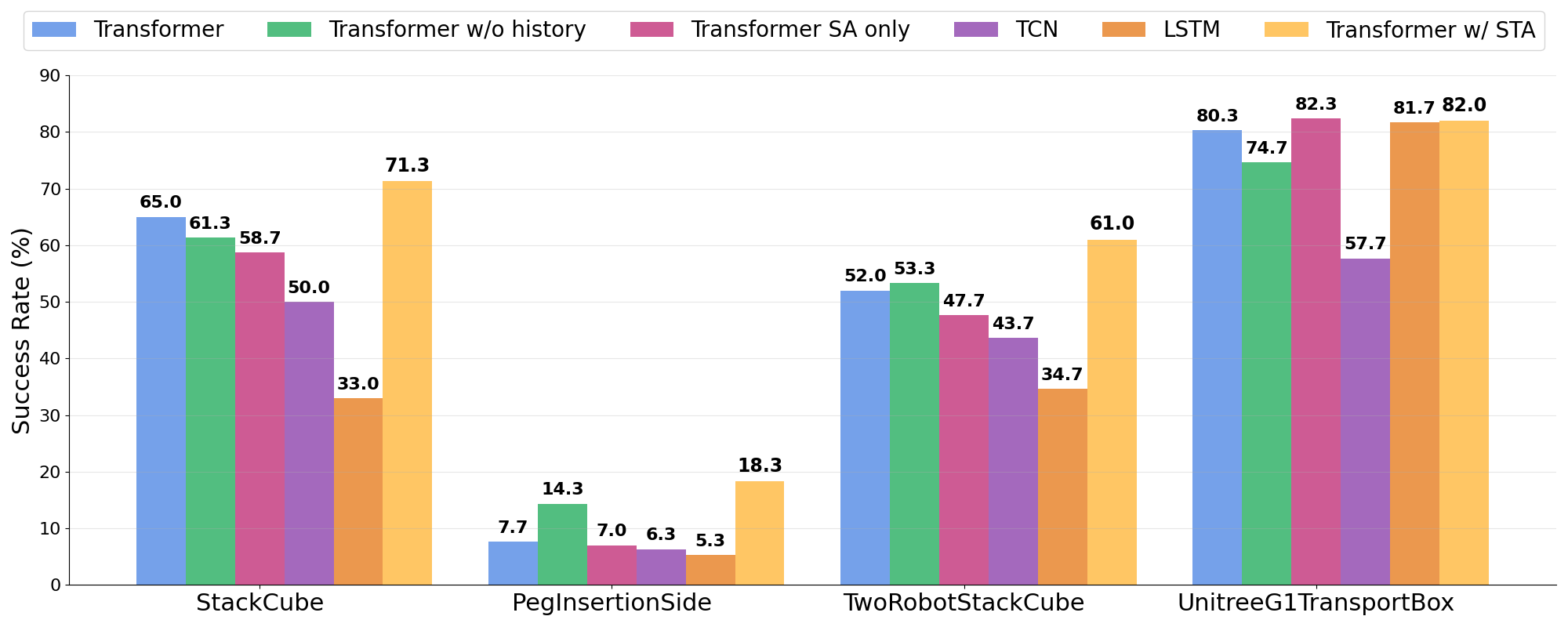}
        \caption{Success rate comparison across four ManiSkill \cite{mu2021maniskill} manipulation tasks. All methods were trained for 50 epochs on recovery-rich demonstrations with periodic validation performed directly in the simulation environment. Results represent the best validation checkpoint performance averaged over 3 seeds with 100 episodes per evaluation. Variance between seeds was negligible and is omitted for clarity.}
      \label{fig:main-results}
   \end{figure}

Fig. \ref{fig:main-results} presents the performance results in terms of success rate, demonstrating the relative effectiveness of different temporal modeling approaches when learning from recovery-rich demonstrations. Our STA Transformer outperforms all baselines across the three precision-requiring and coordination-demanding tasks, with particularly notable improvements on \textit{PegInsertionSide}, reporting more than a 2× improvement over standard Transformer (18.3\% vs 7.7\%). Notably, the Transformer with self-attention only performs similarly to or below its cross-attention variant, while both acting with full history occasionally underperform the no-history baseline, together suggesting that naive integration of historical information does not always benefit policy performance. Traditional sequence modeling approaches show consistent limitations in this domain, with LSTM performing particularly poorly on precision-critical tasks.
On \textit{UnitreeG1TransportBox}, performance is comparably high across all methods. We attribute this to the task's inherent robustness to the noise introduced during data collection: the privileged policy used for demonstration generation recovers reliably without exhibiting meaningful corrective behaviors, limiting the temporal structure present in the training data and consequently reducing the signal available for STA to exploit. These results support our hypothesis that structured attention mechanisms designed for state transition modeling can provide meaningful advantages over history-agnostic and standard sequence modeling approaches, with gains amplified where demonstrations contain rich temporal structure and tasks present challenging execution conditions.

   \begin{figure*}[t]
      \centering
      \includegraphics[scale=0.33]{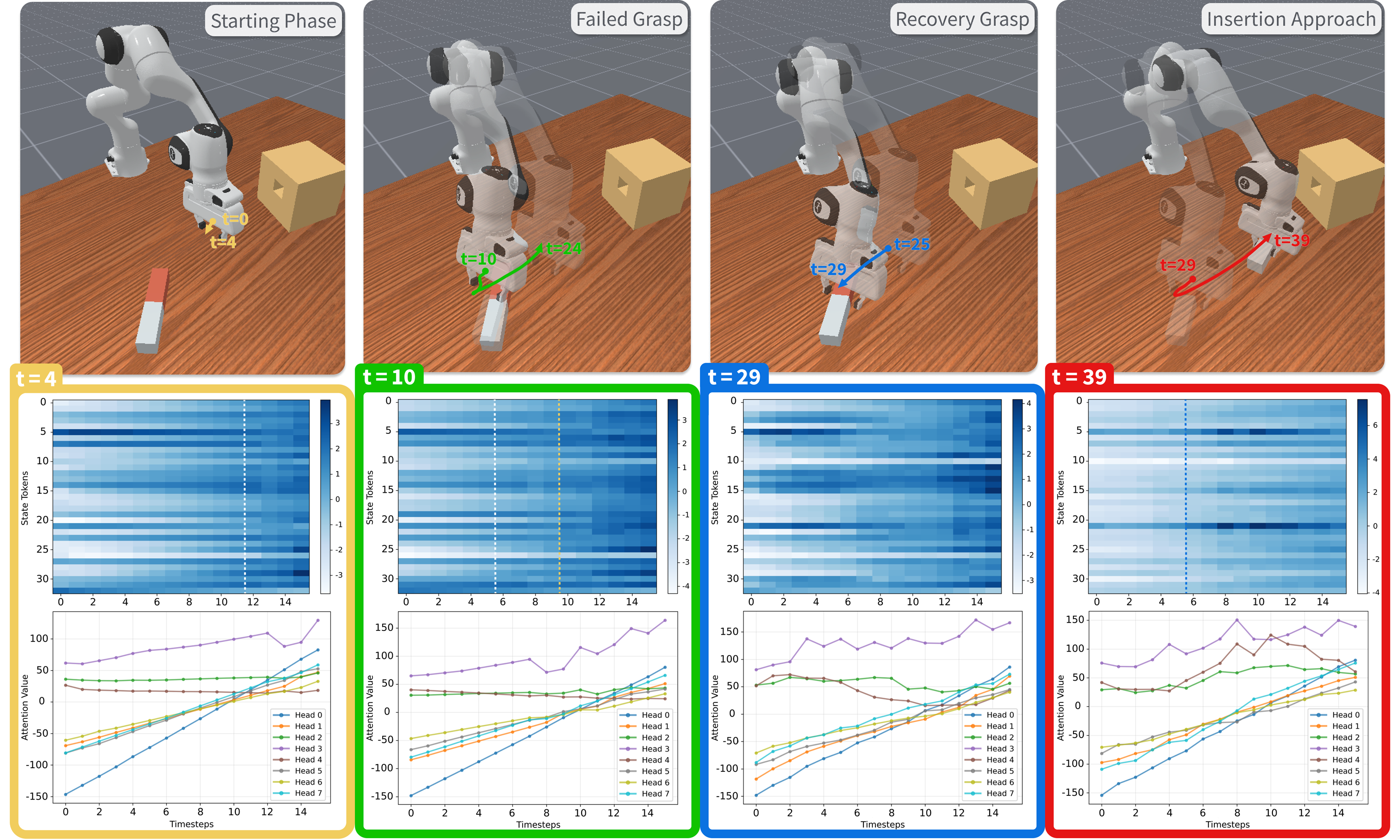}
        \caption{Attention pattern analysis during a \textit{PegInsertionSide} trajectory showing four critical execution phases. \revA{Top heatmaps display state transition weights averaged across heads for individual \textit{state tokens} over a 15-step history window (darker = higher attention). Bottom graphs show the complementary view: per-head attention values summed across all tokens, revealing which heads contribute most to historical retrieval.} Both visualizations share the same temporal axis where timestep 0 represents the oldest historical information and timestep 15 the present. Vertical dotted lines indicate the corresponding timestep in the temporal sequence for each colored execution phase. The visualization demonstrates how STA learns to selectively retrieve relevant historical context during recovery phases (t=29) compared to initial execution patterns (t=4, t=10).}
      \label{fig:attention-analysis}
   \end{figure*}

\begin{table}[b]
\caption{\revA{Networks architecture details.}}
\label{tab:params-networks}
\resizebox{\columnwidth}{!}{%
\begin{tabular}{@{}llll@{}}
\toprule
                         & \textbf{TCN} & \textbf{LSTM} & \textbf{Transformer} \\ \midrule
\textbf{Decoder size}    & 512          & 512           & 512                              \\
\textbf{Encoder size}    & 512          & 512           & 512                              \\
\textbf{N Layers}        & 4             & 4            & 4                    \\
\textbf{N Heads}         & -             & -            & 8                    \\
\textbf{Normalization}   & LayerNorm     & LayerNorm    & RMSNorm              \\
\textbf{FFN Non Linearity}   & ReLU             & ReLU            & GELU                 \\
\textbf{Final FFN}       & \cmark            & \cmark            & \xmark                    \\
\textbf{Batch Size}      & 16          & 16           & 16                   \\
\textbf{Sequence Length} & 16          & 16           & 16                   \\
\textbf{Learning Rate}   & 8e-5        & 8e-5         & 8e-5                   \\
\multirow{3}{*}{\textbf{Parameters}} & \multirow{3}{*}{9.7M} & \multirow{3}{*}{10.4M} & 21.2M (Transformer) \\
                                     &                       &                        & 17M (Transformer SA only) \\
                                     &                       &                        & 23.3M (Transformer w/ STA) \\
\bottomrule
\end{tabular}
}%
\end{table}

\subsection{Attention Pattern Analysis}
Beyond overall performance metrics, we analyze the learned attention patterns to understand how STA processes temporal information. We examine state transition scores $S_{t-k:t}S^T_t$ that modulate historical attention scores $Q_{t-k:t}K^{T}_{t-k:t}$ at timestep $t$. Fig. \ref{fig:attention-analysis} shows a trajectory executed by our transformer-based policy in the \textit{PegInsertionSide} task, which involves reaching the peg, lifting it from the white end, and inserting it into the box hole. We analyze how state transition values evolve across four key execution phases: starting phase (t=4), failed grasping attempt (t=10), successful recovery grasp (t=29), and final insertion approach (t=39). The visualization shows values extracted from the last Transformer layer, which provides the most interpretable results. \revA{The top heatmaps display state transition weights averaged across all heads for \textit{state tokens}, showing their individual relevance across the 15-step history window (darker colors indicate stronger attention). The bottom graphs show state transition values summed across tokens per computation head, revealing which heads contribute the most to historical context retrieval at each timestep.}
Starting from a pre-initialized history of identical starting states -- strategy adopted strictly for analysis purposes -- we observe initial patterns showing lower state transition scores for past states and higher scores for current state information (t=4). This pattern persists through the first grasping attempt (t=10) that ultimately fails. The comparison between the first and second grasping attempts reveals a striking change: during the successful recovery attempt (t=29), the patterns show higher state transition scores extending more into the past, with particular activation of heads 2 and 4 that appear to function as retrieval pathways for relevant historical state relationships. The final trajectory phase (t=39) shows restored focus on recent timesteps with higher state transition scores on specific tokens, likely facilitating the precise movements necessary for successful peg insertion.

This analysis provides insights into how STA learns to selectively attend to relevant past events during challenging execution phases, while downweighting irrelevant historical information in others, supporting our hypothesis that structured attention mechanisms can better exploit the temporal dependencies present in recovery-rich demonstrations.

\subsection{Impact of Temporal Masking on Training and Inference Robustness}

   \begin{figure}[t]
      \centering
        \includegraphics[scale=0.33] {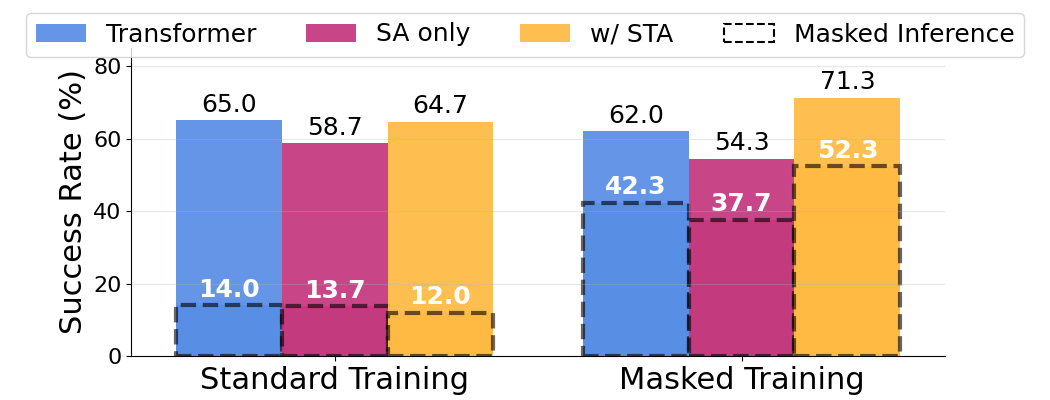}
        \caption{Ablation study comparing temporal masking effects during training and robustness at inference time. Results show the \textit{StackCube} task performance under different training (standard vs. masked) and inference (complete vs. partially masked observations) conditions. Differently from baseline approaches our proposed STA Transformer benefits from temporal masking training, demonstrating architecture-specific advantages.}
      \label{fig:ablation-masking}
   \end{figure}

To understand the specific contributions of our temporal masking training strategy, we evaluate both its direct effects on performance and the robustness it provides at inference time. We compare policies from standard and masked training at inference with both complete and partially masked observations. For 'masked' inference evaluation, we remove visual observations for $n$ consecutive steps where $n$ ranges from 0 to $L$-1, with $L$ being the total sequence length, ensuring at least one timestep retains the full observation, with n resampled with 0.1 probability when it reaches zero, otherwise standard inference conditions apply.
Fig. \ref{fig:ablation-masking} presents results for the \textit{StackCube} task, revealing several important findings. Our STA Transformer trained with masking achieves superior performance under standard inference conditions compared to the same model trained without masking (71.3\% vs 64.7\%), demonstrating that temporal masking enhances learning even when full observations are available during deployment. This suggests that forcing the model to rely on historical context during training develops more robust temporal reasoning capabilities. In contrast, neither of the other Transformer baselines benefits from masked training -- the standard Transformer loses 3.0\% and the SA only variant 4.4\% -- confirming that the effectiveness of temporal masking is specifically tied to our state transition attention mechanism rather than being a general improvement applicable to any architecture. Notably, under masked inference conditions, our STA Transformer maintains a significant advantage over baselines (52.3\% vs 42.3\% and 37.7\%).

\subsection{Historical Context Dependency}
\begin{figure}[t]
      \centering
        \includegraphics[scale=0.17]{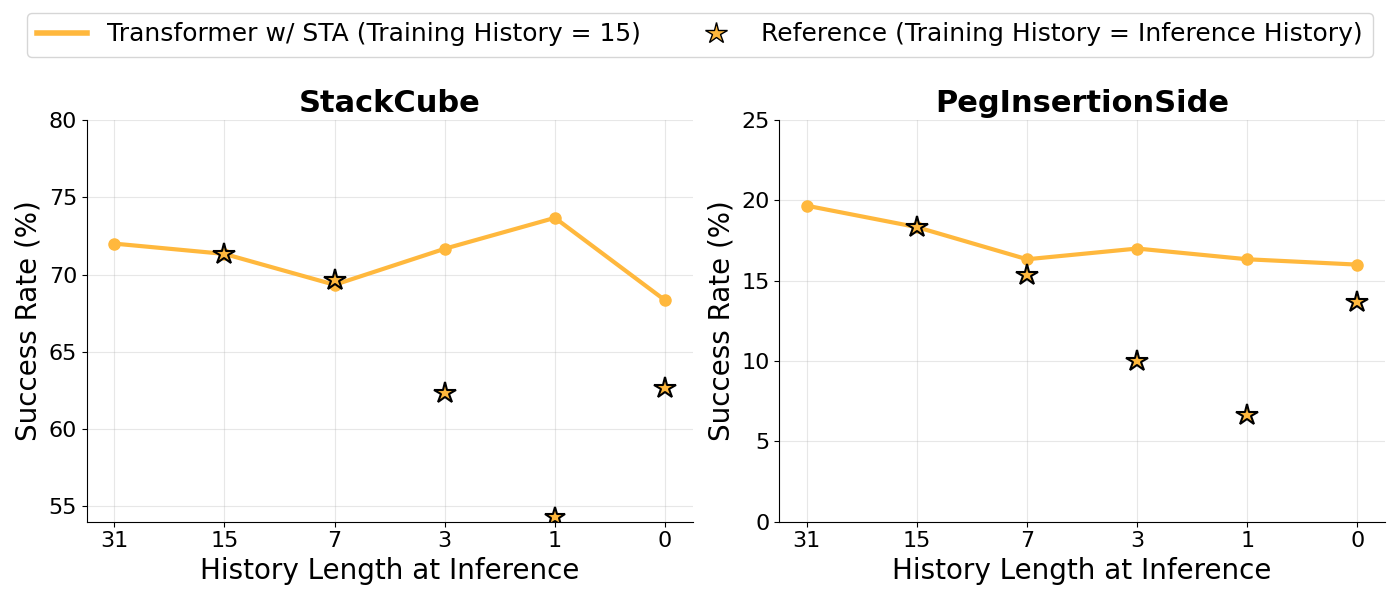}
        \caption{Historical context dependency analysis showing performance vs. inference history length for a STA Transformer trained with 15-step history. Star markers indicate reference performance from models trained with inference-matching history lengths. Results demonstrates robust performance across different inference history lengths, with \textit{StackCube} showing remarkable stability and \textit{PegInsertionSide} exhibiting graceful degradation.
        }
      \label{fig:ablation-history}
   \end{figure}

We analyze how history length at inference affects the performance of our STA Transformer, focusing on \textit{StackCube} and \textit{PegInsertionSide}. We train our method with 15 historical steps (16-step sequences) and evaluate with variable history lengths. Additionally, we train \revA{\textit{reference policies}} -- STA Transformers trained with history lengths of 7, 3, 1, and 0 steps -- to provide baseline performance where training and inference settings match.
Results in Fig. \ref{fig:ablation-history} demonstrate the robustness of our approach when evaluated with limited historical information available. For \textit{PegInsertionSide}, we observe only modest performance degradation as inference history decreases, while in \textit{StackCube}, our method occasionally even exceeds the initial performance. This robustness suggests that our STA mechanism enables learning effective decision-making from rich historical information during training, which transfers well to inference scenarios with truncated historical context. Relative to reference policies trained with shorter histories, we observe substantial performance drops, particularly for those trained with 1 and 3 historical steps. We attribute this degradation to our temporal masking procedure during training, which may excessively dampen the training signal when applied to sequences with already limited temporal information. Notably, the trained \revA{\textit{reference policy}} with 0 historical steps at training (where no temporal masking is applied), presents higher performance than those with 1 and 3 historical steps. This interaction between masking strategy and history length highlights the importance of adequate temporal context for our training approach to be effective.

\section{Discussion}
Our work demonstrates how structured attention mechanisms can effectively exploit temporal dependencies present in recovery-rich demonstrations, as evidenced by attention pattern analysis showing adaptive retrieval of historical information during recovery phases and by consistent performance improvements across precision-requiring and coordination-demanding tasks. STA's advantage emerges specifically in scenarios combining rich temporal structure in training data with challenging execution conditions, while benefits diminish when demonstrations present limited diversity in failure and recovery patterns, as observed in \textit{UnitreeG1TransportBox}.
\revB{We acknowledge limitations constraining the generalizability of our findings. The evaluated tasks are relatively short-horizon and, as evidenced by non-trivial baseline performance without historical context, do not fundamentally require temporal reasoning for basic task completion. Our improvements specifically target the challenging cases where imprecise movements or unrecoverable situations benefit from historical awareness. \revC{More complex temporally-extended tasks with stronger partial observability -- where current observations alone would be insufficient for informed decision-making -- would provide additional validation of historical reasoning capabilities, though extending to such scenarios is primarily limited by hardware requirements for training with extended sequences and storing larger histories at inference}.} \revB{Future extensions should address scalability concerns through memory-efficient techniques for both training and inference phases}. \revB{Our evaluation is conducted entirely in simulation, though no inherent architectural barriers prevent real-world deployment beyond standard sim-to-real transfer challenges common to vision-based policies}.
\revC{Additionally, our data collection methodology remains constrained by the capabilities of privileged policies used to generate failure and recovery patterns, limiting the diversity and complexity of temporal dependencies that can be learned. \revD{Alternative approaches including human demonstrations of natural recovery behaviors or online learning could provide richer temporal structure for future investigation}.}

\section{Conclusions}
We have presented CroSTAta, employing a State Transition Attention mechanism that modulates attention weights based on learned state evolution patterns to improve temporal reasoning in robotic manipulation policies. Our experimental evaluation demonstrates that STA outperforms standard temporal modeling approaches across manipulation tasks, with particularly notable improvements in precision-critical scenarios. The ablation studies reveal additional findings regarding temporal masking and demonstrate robustness to reduced historical context at inference time. These results establish our approach as a promising direction for developing more capable manipulation policies that can effectively learn from and reason about information history.


\bibliographystyle{ieeetr}
\bibliography{sample}

\end{document}